\documentclass[letterpaper, 10 pt, conference]{ieeeconf}
\IEEEoverridecommandlockouts

\usepackage[backend=biber, sorting=none]{biblatex}
\bibliography{references}

\usepackage{amsmath,amssymb,amsfonts}
\usepackage{graphicx}
\usepackage{textcomp}
\usepackage{algorithm}
\usepackage{algpseudocode}
\usepackage{tabularx}
\usepackage{multirow}
\usepackage{soul}

\def\BibTeX{{\rm B\kern-.05em{\sc i\kern-.025em b}\kern-.08em
    T\kern-.1667em\lower.7ex\hbox{E}\kern-.125emX}}

\makeatletter
\newcommand*\titleheader[1]{\gdef\@titleheader{#1}}
\AtBeginDocument{%
  \let\st@red@title\@title
  \def\@title{%
    \bgroup\normalfont\large\centering\@titleheader\par\egroup
    \vskip1.5em\st@red@title}
}
\makeatother

\title{
A Probabilistic Formulation of LiDAR Mapping with Neural Radiance Fields \\
}

\titleheader{\scriptsize This work has been submitted to the IEEE for possible publication. Copyright may be transferred without notice, after which this version may no longer be accessible.}
\author{Matthew McDermott$^{1}$ and Jason Rife$^{2}$
\thanks{$^{1}$Matthew McDermott is a student in the Mechanical Engineering Ph.D. program at Tufts University in Medford, MA. He works in the Automated Systems and Robotics Laboratory (ASAR) with Dr. Jason Rife. He received his B.S. and M.S. degrees in Mechanical Engineering at Tufts University,
        {\tt\small matthew.mcdermott@tufts.edu}}%
\thanks{$^{2}$Jason Rife is a Professor and Chair of the Department of Mechanical Engineering at Tufts University in Medford, Massachusetts. He directs the Automated Systems and Robotics Laboratory (ASAR), which applies theory and experiment to characterize integrity of autonomous vehicle systems. He received his B.S. in Mechanical and Aerospace Engineering from Cornell University and his M.S. and Ph.D. degrees in Mechanical Engineering from Stanford University.
        {\tt\small jason.rife@tufts.edu}}%
}

\begin{document}
\maketitle

\begin{abstract}
In this paper we reexamine the process through which a Neural Radiance Field (NeRF) can be trained to produce novel LiDAR views of a scene.
Unlike image applications where camera pixels integrate light over time, LiDAR pulses arrive at specific times. As such, multiple LiDAR returns are possible for any given detector and the classification of these returns is inherently probabilistic.
Applying a traditional NeRF training routine can result in the network learning ``phantom surfaces'' in free space between conflicting range measurements, similar to how ``floater'' aberrations may be produced by an image model.
We show that by formulating loss as an integral of probability (rather than as an integral of optical density) the network can learn multiple peaks for a given ray, allowing the sampling of first, $\text{n}^{\text{th}}$, or strongest returns from a single output channel. Code is available at \url{https://github.com/mcdermatt/PLINK}
\end{abstract}

\section{Introduction}

Neural Radience Fields (NeRFs) provide continuous representations of scenes by storing information about the surrounding world inside the weights of a neural network \parencite{nerf}. 
Recent works have extended NeRFs from camera images to LiDAR point clouds for use in localization \parencite{pan2024pin}, odometry \parencite{nerfloam}, path planning \parencite{shubPathPlanning}, and data augmentation \parencite{NFL, lidarNeRF}.

To date, LiDAR applications of NeRFs have assumed a deterministic model of the scene. 
In existing approaches, a LiDAR depth map is computed from the network output in the same way as one may be produced from a conventional NeRF, where the weighted sum of opacity at various test points along each ray is used to estimate depth in that direction \parencite{NFL, lidarNeRF, nerflidar}.
Researchers have identified several geometry-based tweaks to enhance the accuracy of LiDAR NeRFs within this deterministic formulation.
For instance, Huang et al. introduced a beam spreading model to enable their algorithm \textit{NFL} to compute range and reflectance accurately for oblique surfaces \parencite{NFL}. Tao et al. take a simpler approach, with their algorithm \textit{LiDAR-NeRF}, masking oblique surfaces to avoid associated complications \parencite{lidarNeRF}.
\textit{NeRF-LOAM} improves sample efficiency by learning an octree map to focus the network on occupied regions of the scene \parencite{nerfloam}. Similarly, \textit{SHINE Mapping} uses a hierarchical spatial structure to encode information and varying levels of detail \parencite{shine}.
Rather than learning a single large network, \textit{PIN-SLAM} achieves a scene representation through many small localized networks in a voxel structure, each relative to a learnable pose parameter, rather than a fixed location in SO(3), which allows for loop closure without retraining the model \parencite{pan2024pin}. 

Despite their impressive accuracy over most segments of a scene, existing algorithms struggle to handle viewpoints from which multiple returns are recieved, since the measured range value is essentially probabilistic. Of the above methods, only \textit{NFL} attempts to address the problem, electing to learn two separate depth channels to represent the first and second returns from any viewpoint \parencite{NFL}. However, \textit{NFL} always assumes exactly two returns, never more and never fewer. 

In this paper, we contend that a small but impactful change to NeRF allows for the direct representation of multiple possible returns, as can occur when a LiDAR views a surface and beyond (as for a window or loose foliage). 
Acknowledging an element of randomness in the training data, we seek to learn a probabilistic representation of depth, rather than a single optimal depth value along each ray emanating from the sensor.
As such, the network is trained to represent the world as a probabilistic density, rather than as a deterministic optical density, as in the case of a traditional NeRF.   
Browning et al. introduced a similar stochastic volumetric world model for LiDAR sample generation \parencite{browning20123d}, 
however, to the best of our knowledge, such a representation has not been combined with NeRF or other differentiable rendering techniques.

The remainder of this paper explains our contribution and demonstrates performance in generating continuous world representations from real LiDAR data. In section II, we concisely but quantitatively define the problems associated with LiDAR data reconstruction. In sections III and IV, we propose a solution and provide implementation details.
In Section V we introduce and discuss the results of our proposed method, which we implement and share in our \textit{Probabilistic LiDAR NeRF Codebase} (PLiNK), which we evaluate across two experiments. The first experiment involves an architectural scene from the \textit{Newer College Dataset} \parencite{newerCollege}, and the second involves a simulated drive through an urban environment in the \textit{Mai City Dataset} \parencite{maicity}.
Finally, in section VI we discuss our results and suggest directions for future work.





\section{Problem Statement}

A LiDAR sensor emits a pulse of near-infared light and uses time of flight to estimate distance to a reflecting surface. 
Consecutive pulses from a static sensor should in theory return the same range measurement every time. In practice, the interplay of semi-transparent surfaces, beam divergence, fluttering of loose foliage, discretization effects, and high sensitivity at sharp corners introduces a level of stochasticity into recorded data \parencite{browning20123d, helios}. This is especially apparent when recording data from a fixed sensor, where range measurements in a given look direction may jump between sequential scans.
Nearly all LiDAR devices allow the selection of a rule to handle cases with multiple returns. For example, the Velodyne VLP-32C can be set to record the first or strongest returns \parencite{velodyne}. The time-trace of the received intensity is inaccessible to the end-user, however \parencite{NFL}.  

\begin{figure}[h]
  \begin{center}
  \includegraphics[width=3.3in]{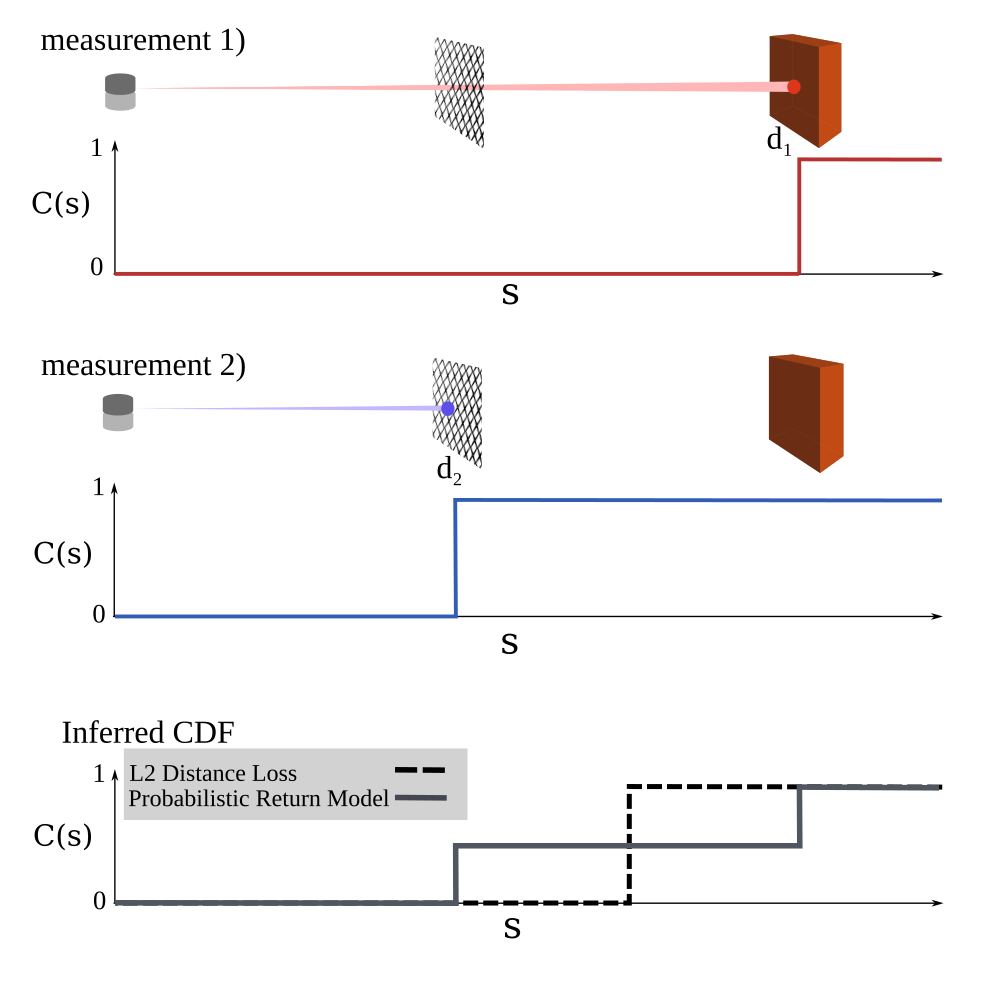}
  \end{center}
  \caption{The stochastic nature of LiDAR returns means that applying a traditional L2 loss on raw distance values will produce phantom surfaces anywhere there is a conflict between two or more measurements. Rather than requiring a single depth estimate to reproduce multiple measurements at different depths, a better approach is to learn an underlying probability density that reproduces the observed measurements.
  }
  \label{fig:CDFFig}
\end{figure}

In order to explain the problem with a deterministic representation of a LiDAR measurement, consider a scene in which a coarse screen is placed in front of a wall, as shown in Fig. \ref{fig:CDFFig}. Along the ray path shown, there is some chance the detector records the measurement as backscatter from the wall (at $d_1$) and some chance as backscatter from the screen (at $d_2$). If the LiDAR records only one or the other of these measurements at a particular instant, then the measurement appears deterministic. Fig. \ref{fig:CDFFig} (top and middle) illustrates these distinct events using a cumulative probability $C(s)$, which describes the probability of backscatter occurring at or before a particular raypath distance $s$.  For a single measurement, the inferred probability jumps from 0 to 1 at the measurement location, implying determinism.  However, if multiple measurements are compiled over time, a more complete picture emerges. For instance, in the case of backscatter occurring from the wall half of the time and from the screen the other half, the cumaltive probability $C(s)$ jumps by 0.5 at each of the two surfaces, as shown by the solid line in Fig. \ref{fig:CDFFig} (bottom).

Because existing LiDAR-based NeRF methods assume deterministic returns, cases of semi-transparency (as shown in Fig. (\ref{fig:CDFFig}) can result in the creation of phantom surfaces. To understand this, consider that a traditional NeRF, determines depth $D$ along a given look direction by  running a forward pass of the network along the ray at test points $t_p \in \mathbb{R}^n$ for each point $p$. This process calculates weighting values $w_p \in \mathbb{R}^n$ associated with each point. These weights are used to construct a weighted depth estimate $D$:

\begin{equation}
    D = \sum_{p = 1}^{n} w_p \|t_p\|_2.
    \label{eq:getD}
\end{equation}

\noindent During training, the loss function guides the NeRF to learn a network that minimizes the L2 Norm between $D$ and available range measurements $d_k$ for each sample $k$, resulting in an optimal depth estimate $D_{\text{NeRF}}^*$:


\begin{equation}
    D_{\text{NeRF}}^* 
    = \arg \min_D \sum_{k=1}^{K} (d_{k} - D)^2.
    \label{eq:NeRFoptimization}
\end{equation}

\noindent This approach creates a phantom surface at $D_{\text{NeRF}}^*$ when working with conflicting LiDAR data, where disparate range returns appear on (or close to) the same sensor ray.
In Fig. \ref{fig:CDFFig}, an L2 loss function sets $D_{\text{NeRF}}^*$ at the halfway point between the near and far measurements, at $D_{\text{NeRF}}^* = \frac{d_1 + d_2}{2}$. 

Phantom surfaces in LiDAR-based NeRFs are the flip side of ``floaters'' in camera-based NeRFs. Like floaters, phantom surfaces are undesirable artifacts that correspond to misplaced real-world surfaces. Floaters occur for camera-based NeRFs due to ambiguity in multi-view reconstruction, when the reconstruction is under-constrained \parencite{MIPnerf360, nerfInTheWild}. By contrast, LiDAR-based NeRFs are highly constrained due to the availability of range measurements, so the opposite problem occurs, with extra measurement data making the system over-constrained. Existing LiDAR-based NeRFs are over-constrained because they assume a single surface is viewable along a raypath; a phantom surface is thus needed to average multiple measurements along a raypath, as illustrated in Fig. \ref{fig:CDFFig} (bottom). Our approach of learning a probabilistic model allows for a consistent representation of data, thereby eliminating phantom surfaces.


\section{Probabilistic Formulation}

\begin{figure}[h]
  \begin{center}
  \includegraphics[width=2.5in]{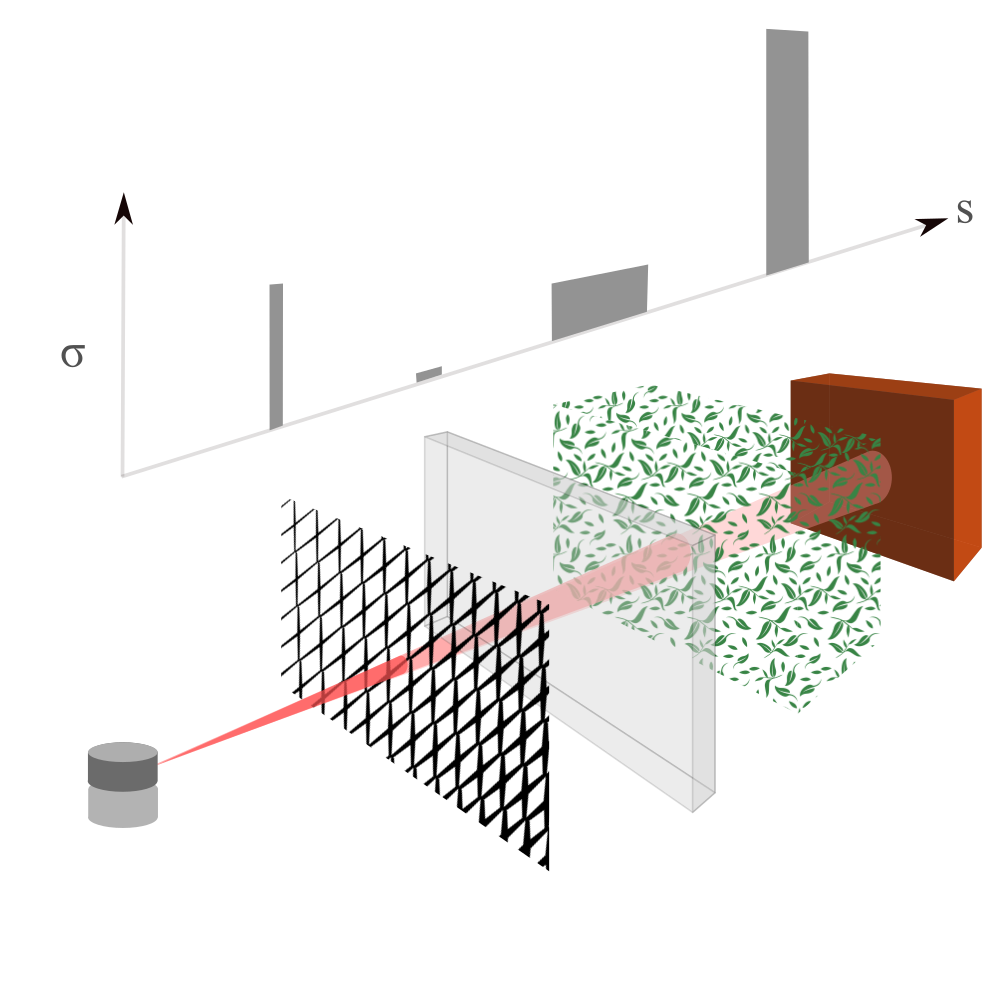}
  \end{center}
  \caption{Visualization of the differential probability  $\sigma$ that a LiDAR return is generated at a path distance $s$ along a ray that passes through a fence, a semi-transparent glass panel, a shrub, and a brick wall.}
  \label{fig:rayFig}
\end{figure}

To model probabilistic LiDAR returns, we replace the optical density field of traditional NeRF algorithms with a field of probability, or more specifically, reflection probability per distance traveled (see Fig. \ref{fig:rayFig}). This differential probability $\sigma(s)$ is integrated along a path coordinate $s$ to create the cumulative distribution shown in Fig. \ref{fig:CDFFig}. The network models the  function $F(\mathbf{x}_0,\mathbf{\hat{\lambda}})$ that provides values of $\sigma$ along $s$ for a LiDAR location $\mathbf{x}_0$ and look direction, as described by the unit pointing vector $\mathbf{\lambda}$. The differential probability $\sigma(s)$ can be obtained by sampling the network.

\begin{equation}
    \sigma(s;\mathbf{x}_0,\mathbf{\hat{\lambda}}) = F(\mathbf{x}_0+s\mathbf{\hat{\lambda}},\mathbf{\hat{\lambda}})
    \label{eq:sigmaFromF}
\end{equation}

Values of $\sigma(s)$ along the raypath can be used to construct $C(s)$, the cumulative probability of a return being generated at or before $s$. Note that $\sigma(s)$ represents the differential probability of a return \textit{given} that the signal arrives at $s$. As such, the conditional $\sigma(s)$ must be multiplied by the prior probability $P(s)$ of successful transmission to $s$. 

\begin{equation}
    \frac{dC}{ds}(s) = \sigma(s) P(s)
    \label{eq:probToS}
\end{equation}

We assume that at each point $s$, the ray is either fully returned (due to reflection or backscatter) or that the ray is fully transmitted. By extension, the cumulative return and transmission probabilities at $s$ are complementary.

\begin{equation}
    C(s) = 1 - P(s)
    \label{eq:complements}
\end{equation}

\noindent Combining (\ref{eq:probToS}) and (\ref{eq:complements}) to eliminate $C(s)$ gives the following.

\begin{equation}
    \frac{dP}{P(s)} = -\sigma(s) ds
    \label{eq:algebra}
\end{equation}

\noindent An integral solution to (\ref{eq:algebra}) can be readily obtained.

\begin{equation}
    \text{ln}(P(s)) = - \int_{0}^{s} \sigma(\gamma) d\gamma
    \label{eq:integral}
\end{equation}

\noindent Rerranging terms and again invoking (\ref{eq:complements}), we obtain the following equation for $C(s)$. 

\begin{equation}
    C(s) = 1 - e^{ - \int_{0}^{s} \sigma(\gamma) d\gamma}
    \label{eq:finalP}
\end{equation}

\noindent Here $\gamma$ is a variable of integration (or \textit{dummy variable}) in the path direction. Importantly, (\ref{eq:finalP}) enforces appropriate constraints for a cumulative distribution, with $C(s)$ guaranteed to increase monotonically from 0 to 1 along the raypath.

Equation (\ref{eq:finalP}) is similar in form, but not meaning, to the physics model originally defined for NeRF \parencite{nerf}. In the original NeRF algorithm, the analogous equation to (\ref{eq:finalP}) computed the intensity at a given camera pixel by integrating the deterministic contributions of various surfaces in the scene (characterized by their optical density). Our formulation replaces that deterministic physics model with a probabilistic one that allows a given return to be generated from various semitransparent surfaces along a ray. A modified loss function is needed to train the network model $F$ as described by (\ref{eq:sigmaFromF}). For the purposes of training the network, we discretize (\ref{eq:finalP}) and perform a numerical integral. The integral is evaluated over a discrete set of path values $\gamma_j$, where in each case the reflectance probability is abbreviated $\sigma_j$.

\begin{equation}
    C(s) =  1 - e^{ - \sum_{j=0}^{J} \sigma_j \delta\gamma_j} = 1 - \prod_{j = 0}^J ( e^{-\sigma_{j} \delta\gamma_j})
    \label{eq:betterC}
\end{equation}

\noindent Here the intervals $\delta\gamma_j$ are the elements of integration, set using the trapezoid rule to be $\delta\gamma_j = \frac{1}{2}(\gamma_{j+1}-\gamma_{j-1})$.

During training, the network model is compared to all range measurements $d_k$ on or near the ray direction under consideration. More specifically, the network is trained by minimizing a loss function $\mathcal{L}_C$ that compares the two-norm of $C(s)$ to a set of unit step functions $u(d_k)$ constructed for each measurement $k \in [1,K]$. 

\begin{equation}
    \mathcal{L}_C = \sum_{k=1}^{K} \int_{0}^{\infty} \big{(}u(d_k) - C(s) \big{)}^2 ds
    \label{eq:ourOptimization}
\end{equation}

\noindent In concept, $C(s)$ could alternatively be estimated as an empirical cumulative distribution function generated along a particular look direction; however, formulation (\ref{eq:ourOptimization}) is a more computationally efficient approach that processes unordered points one-by-one, with no need to sort by raypath.

Once trained, the network can be used to render the scene from a novel viewpoint. Because of the probabilistic construction of the model, the rendering is stochastic. A randomized range sample $D_{render}$ can be obtained, for instance, by transform sampling, a process in which a random sample is drawn from a uniform distribution and used to invert the cumulative distribution $C(s)$ \parencite{inverseTransformSampling}.

\begin{equation}
    D_{\text{render}}(\mathbf{x}_0,\mathbf{\hat{\lambda}}) = C^{-1}(x;\mathbf{x}_0,\mathbf{\hat{\lambda}}), \text{  } x \sim \mathcal{U}(0,1)
    \label{eq:ourDist}
\end{equation}

\begin{figure}[h]
  \begin{center}
  \includegraphics[width=3.5in]{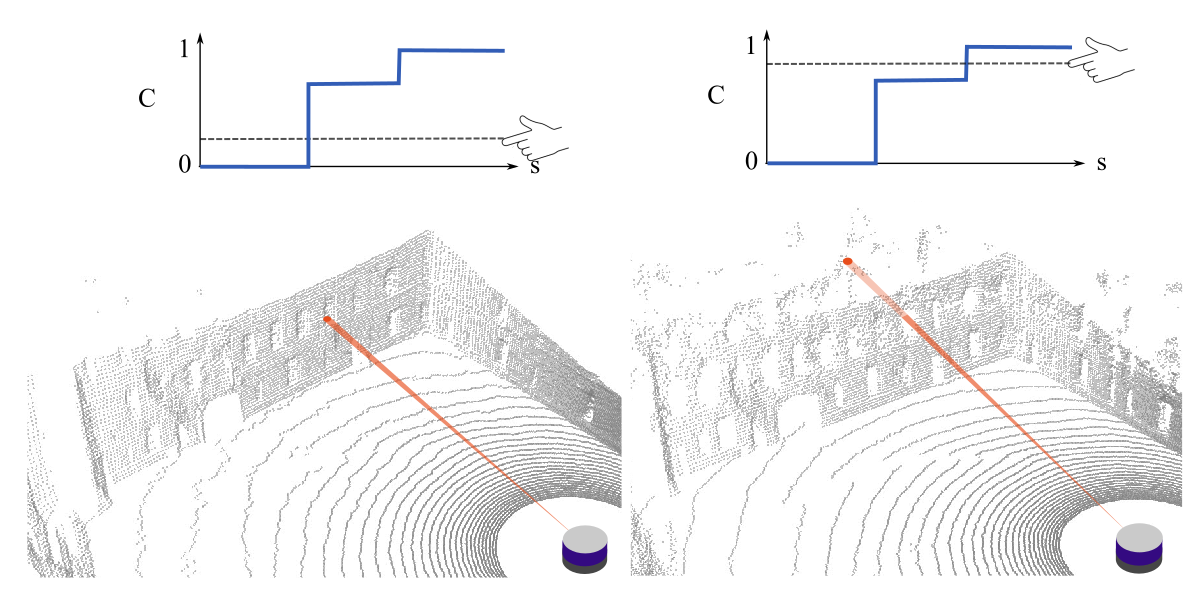}
  \end{center}
  \caption{When multiple returns occur on a raypath, the inverse cumulative distribution can be set to a low value to render nearest returns (left) or a high value to render more distant returns (right). The two renderings are different for directions viewing semi-transparent surfaces (windows) but identical for solid surfaces (walls).}  
  \label{fig:1v95}
\end{figure}

\noindent A deterministic rendering can also be constructed by applying a rule (such as first return, most probable return) or by defining a confidence interval that inverts C(s) for a particular value. An example of the confidence interval approach is shown in Fig. \ref{fig:1v95}. The figure shows two renderings, one at the 10\% confidence interval (left) and another at the 90\% confidence interval (right).

\section{Implementation}

\subsection{Coarse-to-Fine Sampling}

\begin{figure}[h]
  \begin{center}
  \includegraphics[width=2.5in]{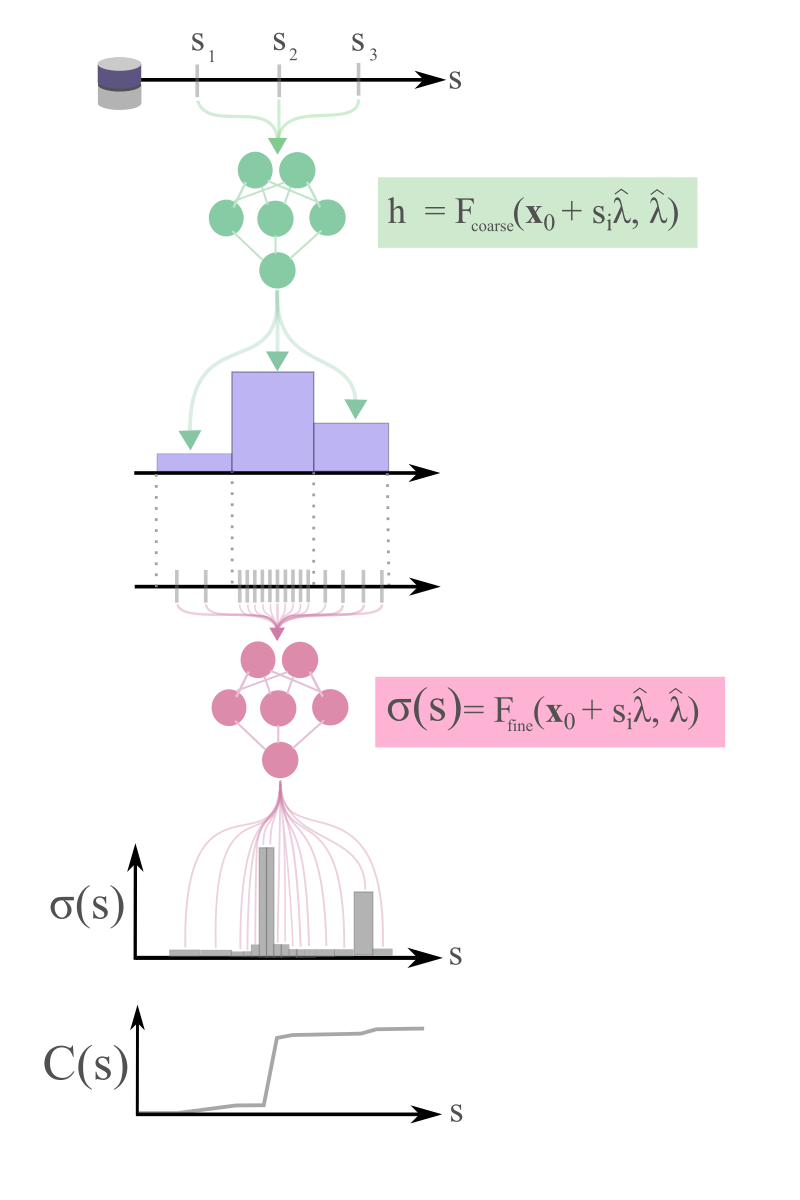}
  \end{center}
  \caption{Outline of a forward training pass through the two networks. Uniformly spaced  points $\mathbf{\gamma}_i$ (histogram bin centers) are passed to the coarse network. The coarse network attempts to estimate reflectance probability within each bin. Normalizing this result produces a histogram (blue) that guides an importance sampling process.  Random draws from the histogram generate a set of test points at which the fine network is evaluated (gray). Loss function (\ref{eq:coarseToFine}) steers the coarse-network histogram to match the fine-network integral within each bin.}
  \label{fig:coarse2fine}
\end{figure}

With the traditional weighted interpolation method of obtaining depth from network output, as described by (\ref{eq:getD}), depth is estimated as a floating point variable.
By contrast, our model builds a reflection histogram over a discrete set of points. Though highly beneficial in avoiding interpolation to phantom surfaces, this discretization can also limit resolution. To achieve high fidelity output with our approach, it is critical to sample as densely as possible around potential surfaces. We thus employ a histogram-based hierarchical sampling method, similar to that of Mip-NeRF 360 \parencite{MIPnerf360}, which focuses sampling in useful regions of the scene and away from free space. This form of hierarchical sampling is sometimes called \textit{coarse-to-fine} sampling.
Our coarse-to-fine approach is summarized in Fig. (\ref{fig:coarse2fine}). 

It is the job of the coarse (or \textit{proposal}) network to estimate a large-scale integral of $\sigma$ over a uniform grid. The coarse network is fed uniform sample locations along a ray, which act as the centers of histogram bins, drawn in blue in Fig. (\ref{fig:coarse2fine}). 
To guide the fine network, trained with (\ref{eq:ourOptimization}), the coarse network implements importance sampling that clusters samples where a surface is likely. Iteratively, the coarse network is optimized to match sample density to the fine network's histogram, using
a second loss $\mathcal{L}_{\text{coarse}}$. The loss $\mathcal{L}_{\text{coarse}}$ considers the $N$ histogram bins along a ray and punishes any bin with height $h_i$ that underestimates the fine-network $\sigma$ integrated over that bin. 

\begin{equation}
    \mathcal{L}_{\text{coarse}} = \sum_{i=1}^{N} \max \Bigg{(} 0,     -h_i(s_i - s_{i-1}) +
    \int\limits_{s_{i-1}}^{s_{i_{\;}}} \sigma(s) ~ d s 
    \Bigg{)}
    \label{eq:coarseToFine}
\end{equation}

\noindent The loss is asymmetric, meaning that it punishes underestimation only. Overestimation is resolved by the normalization of the $\sigma$ integral and $h_i$ sum, as described below. 

\begin{equation}
    \begin{array}{cc}
    \sum\limits_{i=1}^{N} h_i(s_i - s_{i-1})=1, 
    & \int\limits_{s_0}^{s_{K_{\;}}} \sigma(s) ~ ds =1
    \end{array}
    \label{eq:c2fnormalization}
\end{equation}

\noindent Because these quantities are normalized to unity, overestimation in one bin implies underestimation in another.

For the same total neurons, the coarse-to-fine method achieves much higher resolution than the coarse (uniform-grid) network alone, by placing samples only where needed.

\subsection{Ray Drop}

Range measurements are only produced when an emitted LiDAR beam bounces off an object and returns to the sensor. Cases where beams do not return are referred to as \textit{ray drop}, and can occur if an object is outside the maximum range of the sensor or if a reflective object is positioned at an oblique angle. Specific mechanisms must be developed to train NeRFs to reproduce ray drop. Some existing approaches include those of Nakashima et al. \parencite{nakashima2023generative}, who propose using a generative adversarial network (GAN) to predict ray drop in LiDAR data, and those of Zhang et al. \parencite{nerflidar}, who classify regions of non-returns using a U-Net structure \parencite{unet} applied to a cylindrical projection of the LiDAR range measurements.  

In this paper, we adopt a simple approach to classify ray drop by minimizing the binary cross-entropy \parencite{NFL, lidarNeRF} between a ground-truth non-return flag $q_{\text{ray}}$ (with value 0 for ray drop, 1 otherwise) and an additional ray-drop estimate $\hat{q}_{\text{ray}}$ for each raypath. 
Producing this estimate requires an additional network output channel that computes $\hat{\phi_i} \in \mathcal{R}$ at all sample points $s_i$ along a ray. The estimate $\hat{q}_{\text{ray}}$ is obtained by passing the integrated $\hat{\phi_i}$ through a sigmoid function, $S(x)$:

\begin{equation}
    \hat{q}_{\text{ray}} = \text{S} \bigg{(} \sum_i^N \frac{d\text{C}}{d s}_i \hat{\phi}_i \bigg{)}, ~~~~
    \text{S}(x) = \frac{1}{ 1 + e^{-x} }
    \label{eq:poolRayDrop}
\end{equation}

\noindent The result approximates the true ray-drop binary $q_{\text{ray}}$, with $\hat{q}_{\text{ray}} \in (0,1)$. Notably, in (\ref{eq:poolRayDrop}), the contribution of network output $\hat{\phi}_i$ for each point is weighted by the derivative of $C$ at that point's location. This automatically provides a higher weighting to parts of the scene that are in the direct view of the sensor, and suppresses reflection output from occluded regions of the scene.   
The consolidated estimate $\hat{q}_\text{ray}$ is trained for a batch of rays $j\in[1, N]$, where each $j$ represents a particular look direction $\hat{\lambda}$ from a particular LiDAR location $\mathbf{x}_0$. The loss function for that batch is $\mathcal{L}_{\text{drop}}$:

\begin{equation}
    \mathcal{L}_{\text{drop}} = -\frac{1}{N} \sum^{N}_{j=1} \big{[} q_{\text{ray}, j} \log(\hat{q}_{\text{ray}, j}) + (1-q_{\text{ray}, j}) \log(1 - \hat{q}_{\text{ray}, j}) \big{]}
    \label{eq:rayDropLoss}
\end{equation}

\noindent This loss is incorporated into training the fine network by replacing $\mathcal{L}_C$ from (\ref{eq:ourOptimization}) with $\mathcal{L}_{\text{fine}}$.

\begin{equation}
    \mathcal{L}_{\text{fine}} = \alpha \mathcal{L}_C + (1-\alpha) \mathcal{L}_{\text{drop}}
    \label{eq:fineLoss}
\end{equation}

\noindent The new loss function $\mathcal{L}_{\text{fine}}$ is a weighted sum of (\ref{eq:ourOptimization}) and $\mathcal{L}_{\text{drop}}$, where we empirically set the weighting parameter $\alpha$ to a value of 0.999. As in \parencite{MIPnerf360}, the two networks are trained jointly, with the fine network leading and the coarse network following. A stop gradient is placed on $\mathcal{L}_{\text{fine}}$ before calculating gradients for $\mathcal{L}_{\text{coarse}}$ to prevent the fine network from distorting its world representation to more closely align with an incorrect proposal network output. 


A small upside of non-returns is that because there is a maximum distance up to which ranges can be measured, it is easier to guarantee all surfaces in the world fall within a unit cube, as is required to make use of positional encoding. For camera-image data, ambient light may travel over very large distances from background scenery, so clever solutions for compressing distant objects \parencite{MIPnerf360, mega} are needed. Because ranges are limited when working with LiDAR data, we can simply apply a linear scaling, with no spatial compression.

\subsection{Sensor Projection Model}

A mechanically rotating LiDAR reports range measurements for a set of elevation angles, one for each beam, and a set of azimuth angles, associated with the timing of the sample as the rotor spins. These azimuth and elevation angles describe a spherical projection surface, in contrast with the flat image plane (pixel array) of a conventional camera. Previous applications of NeRF have not specifically accounted for the distinction, instead using the default NeRF physics model that assumes a flat image plane.  In this paper, we explicitly reformulate the NeRF physics model to account for spherical projection. 

\begin{figure}[h]
\begin{center}
  \includegraphics[width=2.0in]{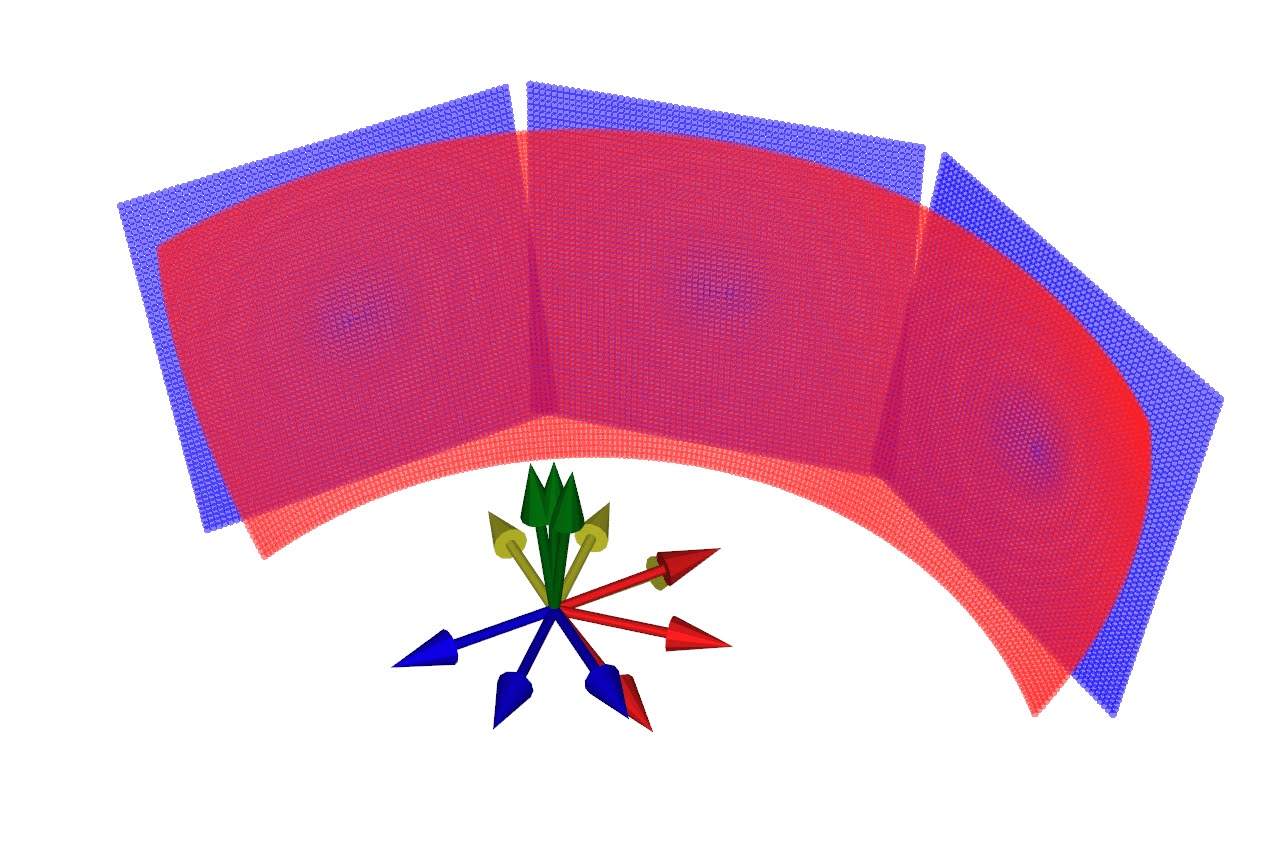}
  \caption{A mechanical LiDAR generates a spherical projection, in contrast with cameras, which generate planar projections. Projection errors can be limited by approximating the spherical projection with many planes. 
  }
  \label{fig:projection}
\end{center}
\end{figure}

In practice, the distinction between spherical and planar projections can be mitigated by dividing the lidar scan into patches, since a spherical projection is well approximated by set of planar projections, as visualized in Fig. \ref{fig:projection}. In the figure, a spherical projection (red) is illustrated as a locus of points at unit range over a span of 25$^\circ$ in elevation and 120$^\circ$ in azimuth. Three planar projections (blue) provide a reasonable approximation,
and errors can be made even smaller with more planar segments. Prior LiDAR-applications of NeRF relied on camera-based pipelines assuming planar projections, so researchers needed to break up the full lidar scan (360$^\circ$) into many smaller patches \parencite{lidarNeRF}. By using a correct (spherical) projection rather than an approximate (planar) projection, we allow for more flexibility in selection of patch size, without introducing projection bias.

Additionally, LiDAR sensors do not have a global shutter and take a finite period of time to record scans. If the sensor is moving during the period of capture, the resulting spherical projection surface is distorted \parencite{VICET}. It is important to update view directions $\lambda$ accordingly, to compensate for the platform pose change between samples within a scan. 

\section{Experiements}

At described in Section II, the \textit{PLiNK} algorithm adds degrees of freedom to the scene model, allowing for more complete representation of semi-transparent features by removing constraints that interpolate among measurements to estimate the most accurate-possible averaged surface along a raypath.  Our modeling decisions thus introduces a tradeoff, prioritizing completeness over accuracy. In this section, we quantify the impact of that trade.

\subsection{Newer College Courtyard}

For our first experiment we benchmark the performance of \textit{PLiNK} on the \textit{Newer College Dataset} \parencite{newerCollege}. We used the ``01 Short Sequence'' trajectory, focusing on a subsequence in which a researcher walks around the perimeter of an outdoor courtyard carrying an Ouster OS1 64 channel LiDAR sensor at chest height. The courtyard is roughly 60m in length and contains a number of windows through which it is possible to see inside the interior of the surrounding buildings. We trained our NeRF using every 5th LiDAR scan 
(with a total of 240 scans in the training set). We removed regions of each scan in which the researcher carrying the LiDAR unit blocked range returns and compensated for motion distortion using \parencite{VICET}, but otherwise did not apply any pre-processing to the raw sensor data. 
Results are compiled in Table (\ref{tab:experiment1Results}) in a form consistent with \parencite{occupancyNetworks, tanksAndTemples}, which quantify performance in terms of Map Completion, Map Accuracy, L1-Chamfer Distance (C-l1), and F-Score. Our algorithm is benchmarked against LiDAR-NeRF \parencite{lidarNeRF}, NeRF-LOAM \parencite{nerfloam}, SHINE Mapping \parencite{shine}, 
and VDBfusion \parencite{vdbfusion}.
In our experiments, both PLiNK and LiDAR-NeRF are implemented with the same network structure, coarse-to-fine sampling process, and training hyperparameters. This was done to isolate the impact of our proposed probabilistic representation on the quality of the learned scene reconstruction. The PLiNK rendering was obtained using three randomized samples along each raypath, to account for the probabilistic model's ability to represent semitransparent surfaces.

\begin{table}[h]
\begin{centering}
\caption{Newer College Dataset}
\label{tab:experiment1Results}
\begin{tabular}{lcccc}
\hline
\textbf{Method}    & \textbf{Completion} $\downarrow$ & \textbf{Accuracy} $\downarrow$ & \textbf{C-l1} $\downarrow$  & \textbf{F-Score} $\uparrow$\\ 
\hline
SHINE     &  14.36  & 8.32  & 11.34  & 90.65\\
VDBfusion & 18.37 & 6.87 & 12.61 & 89.96\\
NeRF-Loam & 15.59  & \textbf{6.86} & 11.24 &  \textbf{91.83}\\
\hline
LiDAR-NeRF & 13.76 & 10.27 & 12.17 & 85.28\\

PLiNK (multi) & \textbf{10.66} & 7.56 & \textbf{9.11} & 90.46 \\ 
\hline
\end{tabular}
\end{centering}
\end{table}

PLiNK achieved the best map completion by significant margin, scoring 10.66 cm, with the next closest methods, LiDAR-NeRF and SHINE, achieving 13.76 cm and 14.36 cm respectively. 
The completion metric describes the quality of the model in minimizing the one-way distance from points in the ground-truth scan to their nearest neighbors in a synthetic scan generated at the same sensor pose. The strong completion score for PLiNK aligns with our hypothesis that the algorithm is particularly adept at modeling extra surfaces where needed, for instance, looking through windows.

By comparison, VDBfusion, and NeRF-LOAM outperform PLiNK in terms of map accuracy. Accuracy describes the opposite one-way distance from points in the synthetic scan to their nearest correspondences in a collocated captured LiDAR scan. This metric quantifies noise in the placement of synthetic world features relative to their true counterparts. Because PLiNK does not constrain noisy measurements to lie on the same surface, we expect some loss of accuracy. However, it is significant to note that SHINE, NeRF-LOAM, and VDBfusion all include hierarchical world subdivision methods (involving multi-resolution voxel grids) to further enhance their accuracy. That PLiNK approaches the accuracy of these methods without the additional complications of a multi-resolution grid suggests that the modeling choices made in PLiNK do not significantly sacrifice accuracy.

The remaining metrics in Table (\ref{tab:experiment1Results}) are values that combine completion and accuracy together. For instance, the L1 Chamfer distance is simply the mean of completion and accuracy. PLiNK's high performance on map completion outweighed its slightly weaker map accuracy, allowing it to obtain the best (lowest) chamfer distance of the tested methods. Similarly, F-Score weight accuracy and completion, by evaluating the harmonic mean between precision and recall metrics, each referring to the percentage of points within a threshold distance of 20cm between ground truth and synthetically generated scans. PLiNK achieves an F-score of 90.46 which is lower but nonetheless close to the leaders,  SHINE (90.65), and NeRF-LOAM (91.83). 

\begin{figure}[h]
\begin{center}
  \includegraphics[width=3.4in]{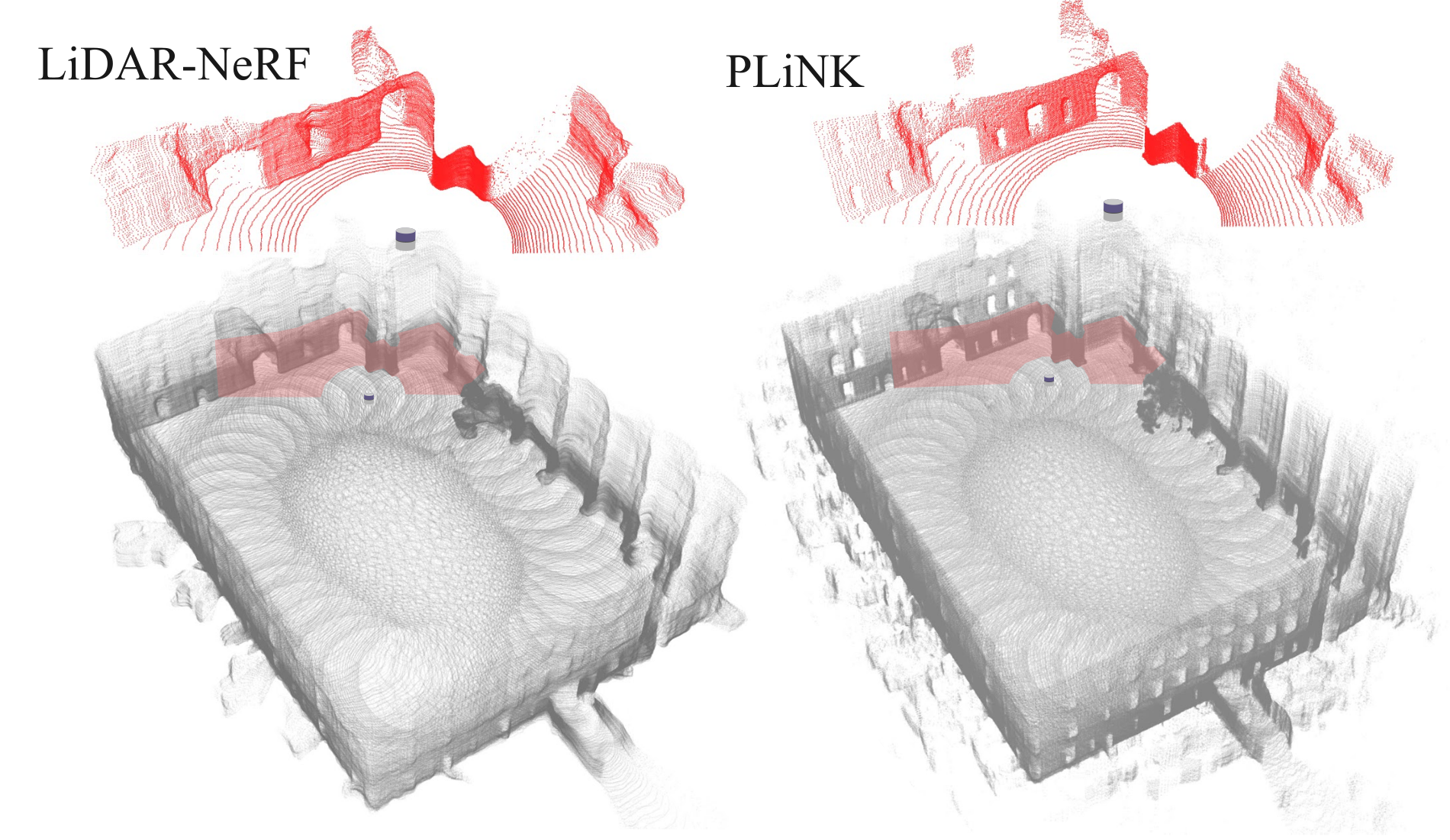}
  \caption{PLiNK resolves windows accurately. Aligned point clouds produced by running LiDAR-NeRF (left) and PLiNK (right) at 40 novel sensor locations are shown. For PLiNK, stochastic samples were drawn from $C$ according to the sampling procedure defined in equation (\ref{eq:ourDist}).  A partial scan at the top of the courtyard is magnified (in red) to highlight the difference in the quality of the learned scene representation between our probabilistic method and that of the deterministic baseline. Unlike PLiNK, LiDAR-NeRF fails to learn partially transparent surfaces and produces lumpy textures and phantom surfaces around windows and sharp corners. }
  \label{fig:plinkvsln}
\end{center}
\end{figure}

A visualization of the reconstructed data helps explain the trends observed in the table. Fig. (\ref{fig:plinkvsln}) shows renderings for both LiDAR-NeRF and PLiNK. LiDAR-NeRF fails to reproduce surfaces on the interior of buildings, instead learning phantom surfaces that protrude from the windows and bubble into the interior of the building. 
This is especially apparent on the series of rooms on left side of the courtyard. Though it is hard to see in the figure, this bubbling effect is not isolated to windows, as many planar surfaces also adopt lumpy aberrations for the LiDAR-NeRF rendering, but not for PLiNK.

\subsection{Mai City}

To emphasize the utility of PLiNK for real-world data analysis, we also consider a synthetic environment, the \textit{Mai City Dataset} \parencite{maicity}. This dataset does not include semitransparent surfaces, so we would not expect PLiNK to offer any advantages over existing methods. 

In this second experiment, we mimic the structure of the first by again comparing performance for the same five methods using the same performance metrics. One change is made in evaluating F-score,
to use a threshold of 10cm to match prior performance analyses that used the Mai City dataset. Because of the faster platform motion (10 $\frac{m}{s}$) more frequent sampling is required than in the \textit{Newer College} experiment in order to not miss narrow alleyways and other briefly visible features. As such, we used even numbered frames as training data and held odd frames for performance evaluation.

The results of the Mai City experiment are shown in Table \ref{tab:experiment2Results}. As expected, PLiNK did not provide a significant completion advantage in this case, since there were never any cases of multiple returns along the same raypath. In this simulated situation, NeRF-LOAM scored best on all performance metrics in the table.  Even with its greatest asset nullified, PLiNK still scored relatively well, similar to NeRF-LOAM and other algorithms in most respects.



\begin{figure}[h]
\begin{center}
  \includegraphics[width=3.in]{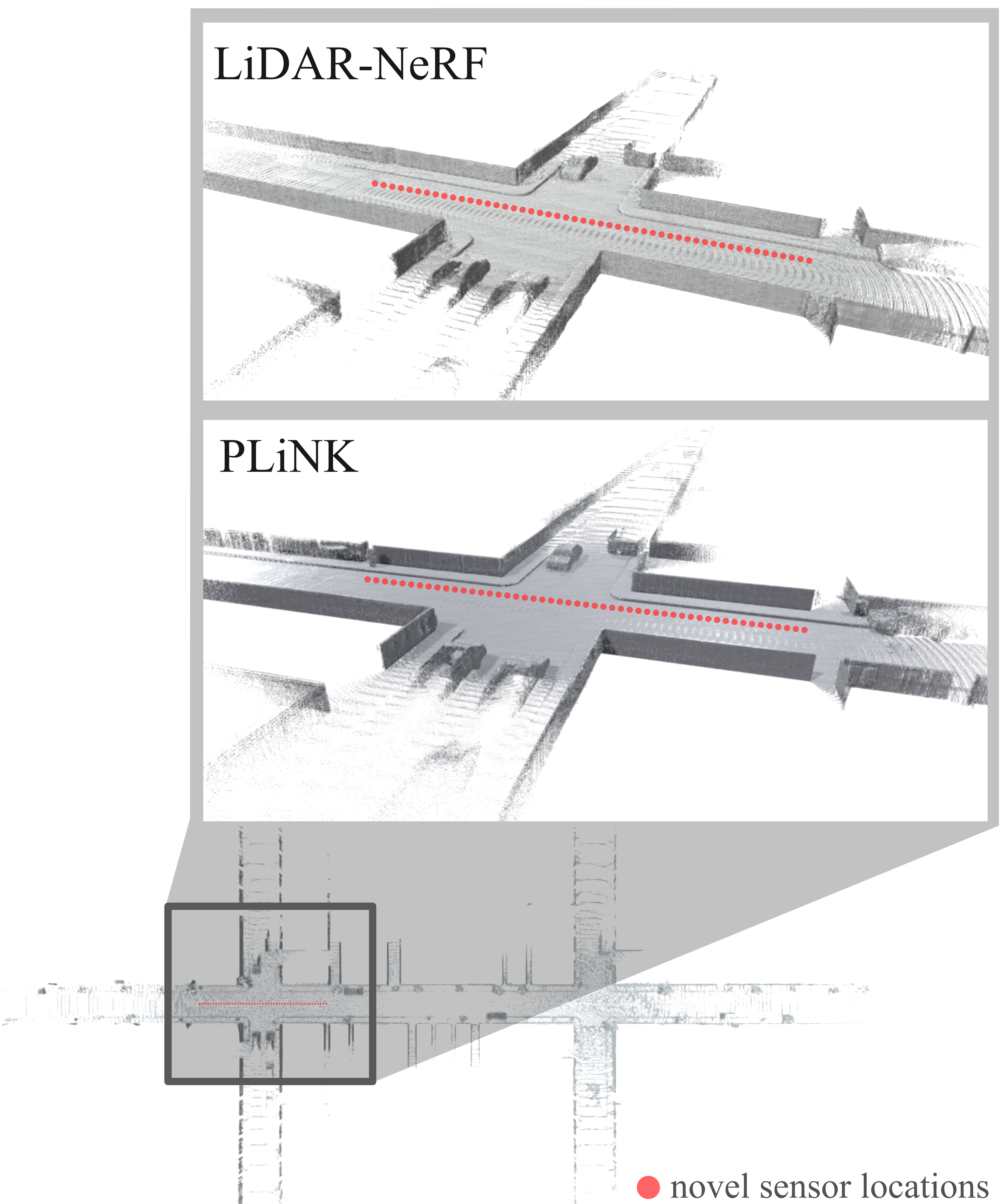}
  \caption{Example of LiDAR-NeRF (top) and PLiNK (bottom) generated point clouds at 50 novel sensor locations in the Mai City dataset. 
  Rendering obtained via cube marching using Open3D \parencite{open3d} 
  }
  \label{fig:MaiCity}
\end{center}
\end{figure}

Again, it is instructive to visualize NeRF renderings, as shown in Fig. \ref{fig:MaiCity}. Renderings for our LiDAR-NeRF and PLiNK implementations are shown. Because conflicting returns on the same raypath are not present for the Mai City simulation, rendered surfaces are much smoother than for rendering real-world data. There there are no phantom surfaces visible for either of methods, and the ripple-like aberrations seen for LiDAR-NeRF in the Newer College scene disappear completely.


\begin{table}[h]
\begin{center}
\caption{Mai City Dataset}
\label{tab:experiment2Results}
\begin{tabular}{lcccc}
\hline
\textbf{Method}  & \textbf{Completion} $\downarrow$ & \textbf{Accuracy} $\downarrow$ & \textbf{C-l1} $\downarrow$ & \textbf{F-Score} $\uparrow$ \\ 
\hline
SHINE     & 5.30 & 4.17 & 4.74 & 89.67\\
Vdbfusion & 8.01 & 4.12 & 6.07 & 90.16\\
NeRF-Loam & \textbf{4.84} & \textbf{3.15} & \textbf{4.00} & \textbf{92.96} \\
\hline
LiDAR-NeRF & 5.65 & 3.81 & 4.73 & 90.16 \\
PLiNK  (Ours) & 5.62 & 3.23 & 4.42 & 90.96 \\ 
\hline
\end{tabular}
\end{center}
\end{table}

\section{Conclusion}

In this paper we proposed a probabilistic approach for training a NeRF from LiDAR data. The results of our evaluation demonstrate how replacing the deterministic depth loss regime with a probabilistic representation can 
allow a network to learn a distribution of range returns along each look direction, rather than forcing the cost function to settle on non-existent phantom surfaces. When combined with a vanilla NeRF backbone, our approach can match or exceed the surface completion performance of existing state-of-the-art methodologies for real world datasets with semitransparent objects. With that said, our approach is still affected by the limitations of traditional NeRF architecture. Training is slow, and real time inference is not possible with our model. We leave the prospect of more efficient hierarchical scene subdivision \parencite{nerfloam, mega, parentChild, takikawa2021neural}, pre-training via graphical primitives \parencite{muller2022instant}, joint learning of sensor poses \parencite{nerfloam, barf}, and potential applications in Gaussian Splatting \parencite{splat} as subjects for future work.

\section{Acknowledgment}

The authors gratefully acknowledge and thank the U.S. Department of Transportation Joint Program Office (ITS JPO) and the Office of the Assistant Secretary for Research and Technology (OST-R) for sponsorship of this work. Opinions discussed here are those of the authors and do not necessarily represent those of the DOT or affiliated agencies.

\printbibliography

\end{document}